\documentclass{article}

\usepackage{arxiv}

\usepackage[utf8]{inputenc}
\usepackage[T1]{fontenc}
\usepackage{hyperref}
\usepackage{url}
\usepackage{booktabs}
\usepackage{amsfonts}
\usepackage{nicefrac}
\usepackage{microtype}
\usepackage{graphicx}
\usepackage{natbib}
\usepackage{doi}
\usepackage{listings}
\usepackage{xcolor}
\usepackage{amsmath}
\usepackage{enumitem}
\usepackage{float}
\usepackage{wrapfig}

\title{An End-to-End Agentic Pipeline for Smart Contract Translation and Quality Evaluation}

\author{
\textbf{Abhinav Goel} \\
Columbia University \\
New York, NY, USA \\
\texttt{ag5252@columbia.edu}
\And
\textbf{Chaitya Shah} \\
Columbia University \\
New York, NY, USA \\
\texttt{cs4621@columbia.edu}
\And
\textbf{Agostino Capponi} \\
Columbia University \\
New York, NY, USA \\
\texttt{ac3827@columbia.edu}
\And
\textbf{Alfio Gliozzo} \\
IBM T.J. Watson Research Center \\
Yorktown Heights, NY, USA \\
\texttt{gliozzo@us.ibm.com}
}

\begin{document}
\maketitle

\begin{abstract}
We present an end-to-end framework for systematic evaluation of LLM-generated smart contracts from natural-language specifications. The system parses contractual text into structured schemas, generates Solidity code, and performs automated quality assessment through compilation and security checks.

Using CrewAI-style agent teams with iterative refinement, the pipeline produces structured artifacts with full provenance metadata. Quality is measured across five dimensions, including functional completeness, variable fidelity, state-machine correctness, business-logic fidelity, and code quality aggregated into composite scores.

The framework supports paired evaluation against ground-truth implementations, quantifying alignment and identifying systematic error modes such as logic omissions and state transition inconsistencies. This provides a reproducible benchmark for empirical research on smart contract synthesis quality and supports extensions to formal verification and compliance checking.
\end{abstract}


\section{Introduction}

Smart contracts, self-executing programs deployed on blockchain networks that automatically enforce agreement terms without intermediaries, have become critical infrastructure for decentralized finance (DeFi), non-fungible tokens (NFTs), and decentralized autonomous organizations (DAOs)~\cite{buterin2014next, wood2014ethereum,capponi2023defi}. However, developing secure and correct smart contracts remains a significant challenge due to the complexity of languages like Solidity, the immutability of deployed contracts, and the high-stakes nature of financial applications~\cite{atzei2017survey, chen2020survey}. 

The gap between domain experts who understand contract requirements and developers who can implement them in Solidity creates bottlenecks in the development process. Traditional approaches require extensive back-and-forth between legal or business stakeholders and blockchain developers, often leading to specification misinterpretation and implementation errors. These errors have resulted in significant financial losses, with notable examples including the DAO hack (\$60M) and the Parity wallet freeze (\$300M)~\cite{luu2016making, mehar2017dao}. Recent work by Anthropic's red team~\cite{anthropic2025scone} demonstrated that frontier AI agents can now autonomously exploit real-world smart contract vulnerabilities, collectively extracting \$4.6M in simulated stolen funds from contracts exploited after model knowledge cutoffs, with exploit revenue doubling every 1.3 months.

Recent advances in Large Language Models (LLMs) have demonstrated remarkable capabilities in code generation~\cite{chen2021evaluating, austin2021program, roziere2024code}. Models such as GPT-4~\cite{openai2024gpt4} and Code Llama~\cite{roziere2024code} can generate syntactically correct code from natural language descriptions. However, generating production-ready smart contracts requires more than syntactic correctness---it demands semantic fidelity to specifications, proper state machine implementation, security best practices, and correct economic logic. Emerging techniques such as SmartInv~\cite{wang2024smartinv} and SmartSys~\cite{wang2024smartsys} leverage multimodal and foundation models to uncover ``machine un-auditable'' smart contract bugs that evade traditional static analysis. These results highlight both the promise and the urgency of developing systematic quality evaluation frameworks for AI-generated smart contracts, given AI's dual-use role in both exploitation and defense.

In this paper, we present an end-to-end framework that unifies generation, verification, and comparative evaluation into a single pipeline, providing a systematic basis for empirical research on smart contract synthesis quality. 

\begin{figure}[H]
    \centering
    \fbox{\includegraphics[width=1\textwidth]{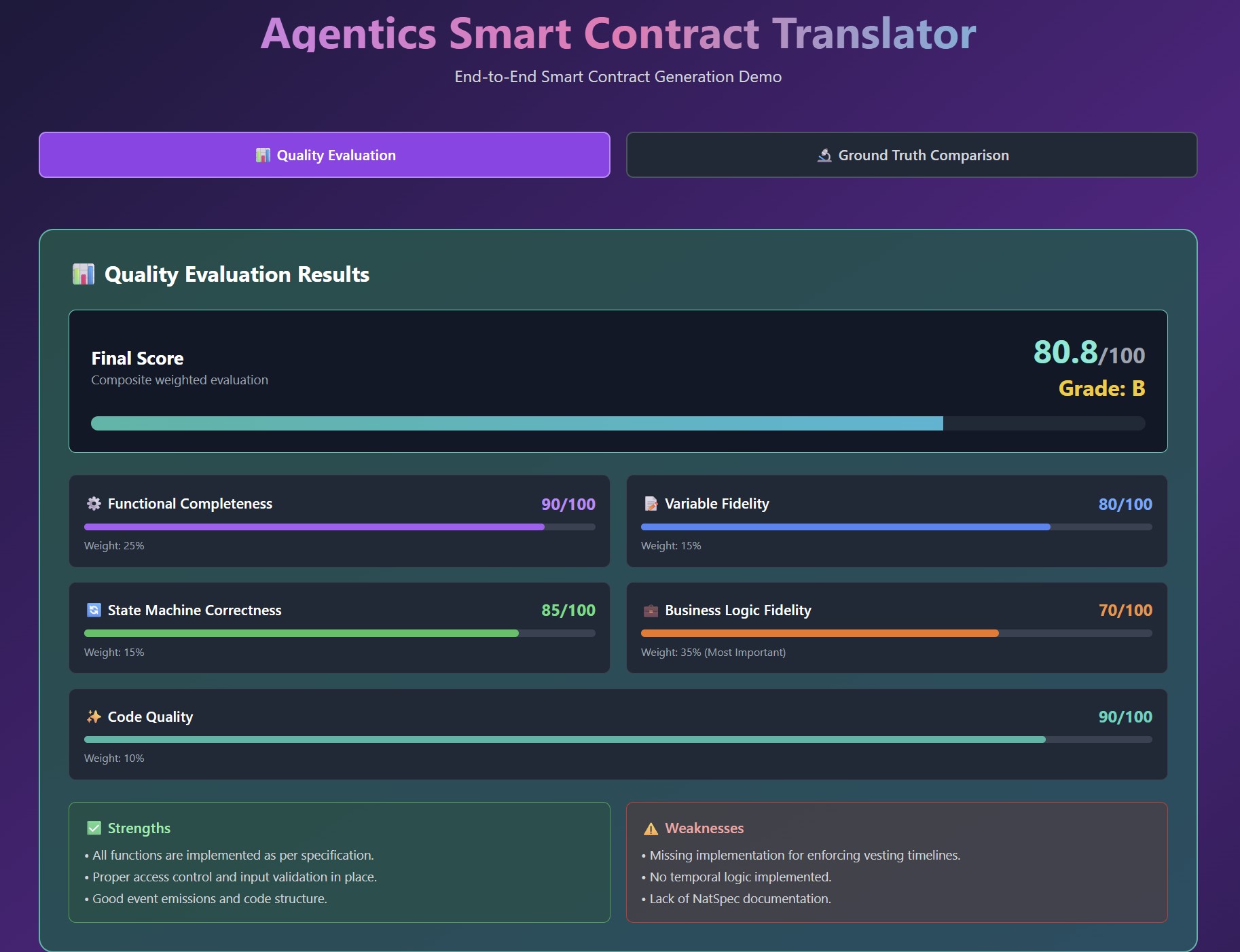}}
    \caption{Single Contract Mode Interface}
    \label{fig:single_contract}
\end{figure}

Beyond batch processing, the system offers an interactive single-contract mode (shown in Figure~\ref{fig:single_contract}) that provides a detailed breakdown of the generation and evaluation process. By highlighting concrete successes (e.g., complete function implementation and proper access control) and pinpointing specific deficiencies (e.g., missing vesting enforcement and NatSpec documentation), this transparent interface enables developers to rapidly identify issues requiring manual review, making the system practical for real-world workflows where human oversight is essential.

Our contributions include:

\begin{enumerate}
    \item \textbf{Multi-Stage Agentic Pipeline}: A comprehensive pipeline orchestrated by CrewAI-style agent teams that (i) parses contractual text into a structured schema, (ii) generates Solidity code and ABI interfaces, (iii) produces optional MCP server wrappers for deployment-adjacent interaction, and (iv) performs automated quality assessment with compilation and security gate checks.
    
    \item \textbf{Five-Dimensional Quality Evaluation Framework}: A rubric grounded in functional and semantic fidelity measuring Functional Completeness (25\%), Variable Fidelity (15\%), State Machine Correctness (15\%), Business Logic Fidelity (35\%), and Code Quality (10\%), with deterministic re-computation of composite scores eliminating drift from model-generated aggregates.
    
    \item \textbf{Agentic Reinforcement Loop}: An iterative refinement mechanism that automatically refines outputs until constraints are satisfied or a fixed iteration limit is reached, with an audit-approval checkpoint mediating security review before downstream artifact release.
    
    \item \textbf{Paired Evaluation Against Ground Truth}: Systematic comparison of generated contracts against expert implementations using the same metric schema, producing per-metric deltas and overall score differences that quantify where generative models align with or deviate from ground truth, enabling targeted analysis of error modes such as logic omissions, state transition inconsistencies, or specification-to-code misalignments.
    
    \item \textbf{Reproducible Benchmarking Infrastructure}: Each stage emits structured artifacts and provenance metadata via streaming events, enabling reproducible analysis, traceability across the generation lifecycle, and support for future extensions including property-based testing, formal verification hooks, and domain-specific compliance checks.

    \item \textbf{Dataset and Code Release}: We release the source code used for smart contract generation, validation, evaluation, and interactive use, as well as an extended version of the corpus which includes automatically generated contracts and their analysis. 
\end{enumerate}

Reliable smart contract generation has direct implications for safety-critical financial infrastructure, legal automation, and compliance-sensitive applications. By standardizing evaluation with transparent metrics, grounded comparisons to expert implementations, and verifiable compilation checks, this framework reduces the risk of silently deploying defective contracts and enables measurable progress in model capability.

\section{Background}

Smart contracts are programs stored on blockchain networks that execute automatically when predetermined conditions are met~\cite{nakamoto2008bitcoin, buterin2014next}. Ethereum, the first and currently most widely used blockchain platform which supports smart contract platform, introduced Solidity as its primary programming language~\cite{wood2014ethereum, solidity2024}.

Solidity is a statically-typed, contract-oriented language designed specifically for implementing smart contracts. Key features include:

\begin{itemize}
    \item \textbf{State Variables}: Persistent storage on the blockchain that maintains contract state across transactions.
    \item \textbf{Functions}: Callable units of code with visibility modifiers (public, private, external, internal) and state mutability specifiers (view, pure, payable).
    \item \textbf{Modifiers}: Reusable code blocks for access control and precondition checking.
    \item \textbf{Events}: Logging mechanism for off-chain applications to track contract activity.
    \item \textbf{Inheritance}: Support for contract composition and code reuse.
\end{itemize}

The immutability of deployed smart contracts makes correctness paramount. Bugs cannot be patched post-deployment without complex migration procedures or proxy patterns~\cite{zheng2020overview}.

Smart contract development demands extreme attention to security because vulnerabilities can lead to catastrophic and irreversible financial losses~\cite{atzei2017survey, chen2020survey}. Common vulnerability classes include reentrancy attacks (where malicious contracts recursively call vulnerable functions before state updates complete), integer overflow conditions, access control flaws allowing unauthorized operations, and timestamp manipulation by miners. Traditional static analysis tools like Slither~\cite{slither2019} and Mythril~\cite{mythril2018} can automatically identify many of these vulnerability patterns, but these tools produce raw findings that require significant security expertise to interpret, prioritize, and remediate.

The emergence of Large Language Models (LLMs) for code generation has opened promising new avenues for automating complex software development tasks~\cite{brown2020gpt3, chen2021evaluating}. Models such as Codex, GPT-4~\cite{openai2024gpt4}, and Code Llama~\cite{roziere2024code} have demonstrated remarkable capabilities in translating natural language descriptions into functional code across various programming domains. However, applying these models to smart contract generation presents unique challenges beyond general-purpose code synthesis. Blockchain-specific terminology carries precise implementation requirements: a ``token'' must conform to established standards like ERC20, ``delegation'' implies particular patterns for proxy contracts and governance rights, and ``staking'' requires careful handling of asset locks and reward distributions. Additionally, many contracts implement explicit state machines that must enforce valid transitions, and economic invariants must be rigorously preserved to prevent value loss or exploitation.

All these challenges makes smart contract generation an ideal testbed for multi-agent AI systems~\cite{yao2023react, wei2022chain}. Rather than relying on a monolithic model to handle all aspects of contract development, frameworks like CrewAI~\cite{crewai2024} and IBM Agentics~\cite{ibm2024agentics} enable the orchestration of specialized agents with distinct roles and capabilities. Individual agents can focus on specific subtasks such as parsing specifications into formal requirements, generating contract code, or auditing for security vulnerabilities. This separation prevents optimistic bias and enables iterative refinement as agents provide feedback to earlier pipeline stages.

\begin{wrapfigure}[37]{r}{0.35\textwidth}
    \centering
    \vspace{-10pt}
    \fbox{\includegraphics[width=0.32\textwidth]{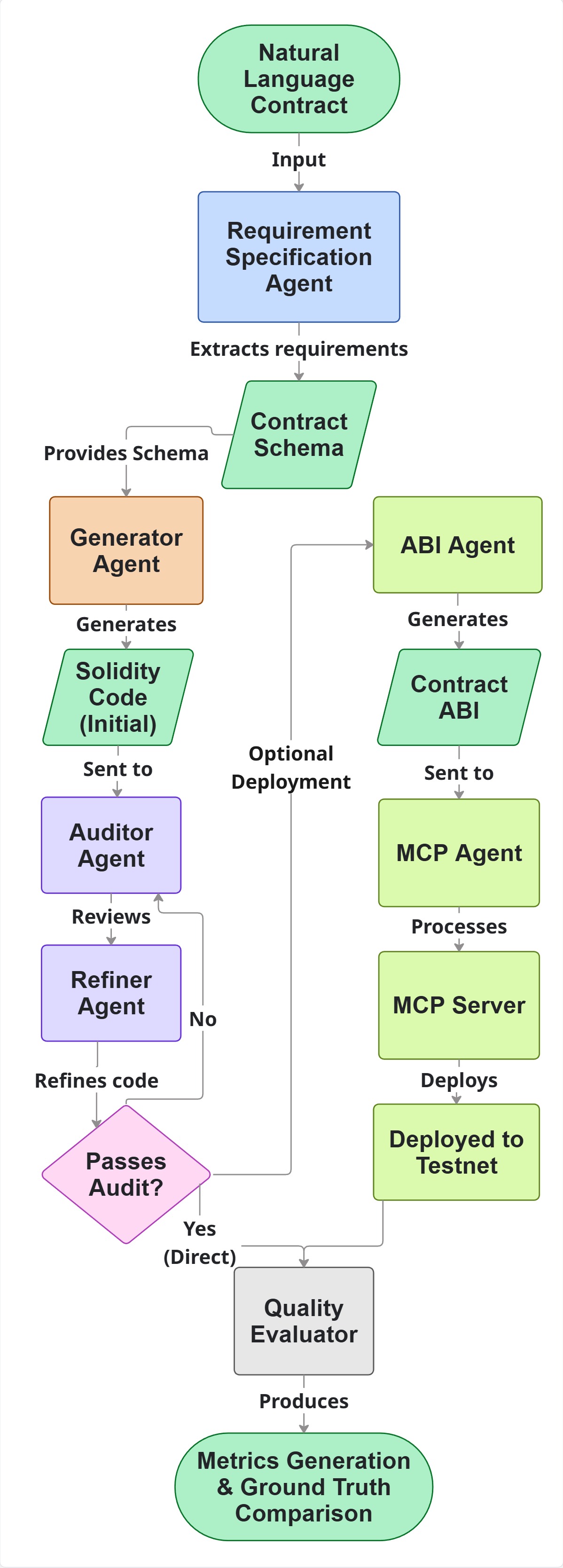}}
    \caption{End-to-end agentic pipeline architecture for smart contract generation, testing and deployment.}
    \label{fig:architecture}
    \vspace{-15pt}
\end{wrapfigure}

To provide rigorous validation beyond what static analysis tools offer, we leverage Finite State Machine (FSM) representations~\cite{fsm2020smart} as formal specifications of intended behavior. FSMs explicitly model contract states (discrete phases in the lifecycle), valid transitions between those states, guard conditions that must hold for transitions to occur, and actions executed during state changes. By comparing a generated contract's actual behavior against its FSM specification, we can verify that the implementation correctly realizes the intended logic—critical when deploying AI-generated code in high-stakes financial environments where post-deployment bugs cannot be easily fixed.

The integration of LLM-powered code generation, multi-agent orchestration, automated static security analysis, and formal verification through FSMs represents a novel approach to safe, reliable smart contract development at scale. The engineering decisions required to orchestrate multiple AI agents effectively, combined with the demonstrated ability to deploy these systems for real-world contract generation, constitute the primary contributions of this applied data science work.

\section{Methodology}

Our system implements a seven-phase agentic pipeline orchestrated by the \texttt{IBMAgenticContractTranslator} class. Figure~\ref{fig:architecture} illustrates the overall architecture.

\subsection{Phase 1: Requirement Specification Agent}

The Requirement Specification Agent extracts structured information from natural language contract specifications into a \texttt{UniversalContractSchema}. The Requirement Specification Agent extracts structured information from natural language contract specifications into a \texttt{UniversalContractSchema} that captures essential contract elements. The schema organizes parties, financial terms, and temporal information (contract dates, deadlines, and time-based conditions). It also structures assets, obligations (party responsibilities), conditions (function names, variable names, state names, transitions, events, and logic conditions), and termination criteria. The agent is instructed to extract \textit{exact} terminology from the specification, avoiding generic placeholders that would lose domain-specific semantics.

\subsection{Phase 2: Solidity Generation}

The Generator Agent transforms the parsed schema into production-ready Solidity code through a structured eight-phase approach. Before generating code, the agent performs semantic analysis to identify the contract's core purpose, workflow structure, state transitions, and economic flows. The generation process enforces 12 critical rules ensuring that specifications translate into enforceable on-chain logic, domain-specific semantics are fully implemented (e.g., ``Token'' requires complete ERC20 compliance), state machines are explicitly enforced, and economic logic is complete and conservative. The agent also maintains a set of forbidden patterns, explicitly avoiding common pitfalls such as empty function bodies, unused variables, silent failures, and decorative code that lacks actual functionality, ensuring all generated code serves a concrete purpose in the contract's operation.

\subsection{Phase 3: Security Auditing}

The Auditor Agent takes the generated Solidity code from Phase 2 and uses an LLM call to perform systematic security analysis across eight critical vulnerability categories: reentrancy attacks, access control flaws, arithmetic safety issues, ether handling vulnerabilities, denial-of-service risks, input validation gaps, timestamp dependence, and external call safety. The LLM analyzes the contract code against these vulnerability patterns and produces a structured audit report with severity levels (none, low, medium, high, critical), specific issues with line references, and concrete remediation recommendations that can be used to refine the generated contract.

\subsection{Phase 4: Reinforcement Loop}

A key innovation in our pipeline is the automated reinforcement loop that iteratively refines generated contracts based on security findings. When the audit identifies medium or higher severity issues, the Refiner Agent automatically remediates vulnerabilities without manual intervention, creating a self-improving feedback cycle that mirrors professional smart contract review while operating autonomously.

\begin{lstlisting}[language=Python, caption={Refinement Decision Logic}]
def should_refine(audit_report, refinement_count, max_iterations=2):
    if refinement_count >= max_iterations:
        return False
        
    severity = audit_report.get('severity_level')
    approved = audit_report.get('approved', False)
    
    if (not approved and severity in ['medium', 'high', 'critical']):
        return True
    return False
\end{lstlisting}

The process enforces a maximum of two refinement iterations to prevent infinite loops while allowing meaningful remediation. The Refiner Agent receives the current contract and detailed audit findings, then applies established secure coding patterns—such as Checks-Effects-Interactions, reentrancy guards, strict access control, and comprehensive input validation—to produce corrected code. Separating generation and refinement agents avoids optimistic bias, and each iteration is re-audited, forming a feedback loop that progressively improves security until the contract passes audit or the iteration limit is reached. This automated reinforcement is especially valuable for smart contracts, where deployment immutability makes rigorous pre-deployment security essential.

\subsection{Phase 5 \& 6: Optional Deployment}

Our pipeline includes the capability to deploy generated contracts to a testnet environment, though this represents an optional branch rather than a required step in the core workflow. When deployment is desired, two additional agents facilitate blockchain interaction. The ABI Agent first generates the Application Binary Interface specification from the refined Solidity code, extracting all contract elements including the constructor with parameters, public/external functions with their inputs, outputs, and state mutability specifications, and events with parameter types and indexing information. These ABI elements are then fed into the MCP Agent, which uses them to generate a complete Model Context Protocol server that exposes all contract endpoints and tools as standardized interfaces. This MCP server enables AI assistants and automation tools to interact with the deployed contract on a simulated Ganache blockchain environment, providing a complete end-to-end workflow from natural language specification to deployable, interactive smart contract. While this deployment capability demonstrates the full potential of the pipeline, the core value proposition—automated generation and security auditing of smart contracts—does not depend on actual deployment and can be leveraged independently for contract prototyping, security analysis, or code generation workflows.

\section{Experimental Setup}

\subsection{Benchmarks}

We utilize the FSM-SCG dataset~\cite{luo2025fsmscg}, a comprehensive collection of 21,976 smart contract specifications specifically designed for evaluating LLM-based contract generation systems. The dataset is publicly available and provides a structured benchmark for end-to-end contract synthesis evaluation. Each entry contains three aligned components: 

\begin{itemize}
\item \textbf{Natural Language Requirements}: High-level descriptions of contract purpose, stakeholder roles, business logic, and constraints (avg. 120 words).
\item \textbf{Finite State Machine (FSM) Specifications}: Formal definitions of states, valid transitions with conditions, guard predicates, and state-change actions.
\item \textbf{Ground Truth Solidity Implementations}: Expert-written contracts (avg. 85 lines) implementing the requirements and FSMs, used as gold-standard references for paired evaluation.
\end{itemize}

The dataset spans Solidity versions 0.4.x to 0.8.x, capturing the language's evolution, including major security improvements (e.g., automatic overflow checking in 0.8.0+), syntax changes (explicit visibility requirements), and modern patterns (immutable variables, custom errors). Contract types include token standards (ERC20, ERC721), crowdsale and ICO mechanisms, governance and voting systems, staking and reward distribution, escrow and payment handling, and access control with multi-signature schemata.\footnote{While complementary datasets exist (e.g., a comment–code dataset with 9,635 function-level snippets), our evaluation focuses exclusively on the FSM-SCG dataset, as it uniquely provides complete requirement–FSM–code triples for end-to-end pipeline evaluation and comparison against ground truth implementations.}

Table~\ref{tab:dataset} summarizes the key characteristics of the FSM-SCG dataset. 

\begin{wraptable}[12]{r}{0.55\textwidth}
\centering
    \vspace{-20pt}
\caption{The FSM-SCG dataset~\cite{luo2025fsmscg}}
\label{tab:dataset}
\begin{tabular}{@{}lr@{}}
\toprule
\textbf{Characteristic} & \textbf{FSM-SCG Dataset} \\
\midrule
Total entries & 21,976 \\
Avg. requirement length & 120 words \\
Avg. contract length & 85 lines \\
Solidity version range & 0.4.x--0.8.x \\
Contract scope & Full contracts with FSM specs \\
Standard tokens (ERC20/721) & $\sim$42\% \\
Custom business logic & $\sim$58\% \\
Components per entry & Requirement + FSM + Code \\
\bottomrule
\end{tabular}
\end{wraptable}

The dataset shows wide variation in contract complexity, from simple token transfers to multi-stakeholder governance systems. The 42\%/58\% split between standard tokens and custom business logic mirrors real-world Ethereum mainnet usage. Coverage across Solidity versions 0.4.x–0.8.x stresses the system's ability to produce modern code, including explicit visibility (0.5.0+), immutable variables (0.6.5+), automatic overflow checking (0.8.0+), and custom errors (0.8.4+). Its requirement–FSM–code structure enables evaluation of natural language parsing, state-machine correctness, and paired comparison against ground truth implementations.

\subsection{Quality Evaluation Framework}

The quality evaluation framework assesses generated Solidity smart contracts across five weighted dimensions. Functional completeness (25\%) verifies that all specified functions are properly implemented with correct naming and complete logic. Variable and parameter fidelity (15\%) ensures the contract maintains naming consistency with the specification and uses appropriate data types throughout. State machine correctness (15\%) evaluates whether the finite state machine is properly defined with valid transitions and guards. The most critical dimension, business logic fidelity (35\%), measures how accurately the contract implements the core semantic requirements including obligations, financial logic, temporal constraints, and conditional logic. Finally, code quality (10\%) assesses implementation craftsmanship by checking for placeholders, error handling, and overall structure. These metrics combine into a weighted composite score that translates to letter grades from A (90-100) to F (below 60), providing a holistic assessment of contract quality.

The composite score is calculated as:
\begin{equation}
    \text{Score} = 0.25M_1 + 0.15M_2 + 0.15M_3 + 0.35M_4 + 0.10M_5
\end{equation}

\subsection{Experiments}

We evaluated the pipeline on 9,000 unique contracts from the requirement\_code.jsonl dataset, processed in six parallel batches of 1,500 contracts each. Every contract passed through the full seven-stage pipeline, excluding deployment-related steps, as the focus was on generation quality and ground-truth comparison. The evaluation used GPT-4o-mini for all agent tasks, allowed up to two refinement cycles, validated buildability with the Solidity 0.8.x compiler, and applied a five-dimensional quality rubric. All six batches produced highly consistent results with minimal variance across metrics, demonstrating strong reproducibility and robustness. The system supports both large-scale \textbf{batch processing} for thousands of contracts and an interactive \textbf{single-contract mode} for detailed, real-time analysis.

\subsection{Code and Data Availability}

To ensure reproducibility and enable independent verification of our results, we publicly release the complete implementation of our agentic pipeline along with all evaluation datasets. The release includes the full source code for all seven pipeline phases (requirement specification, Solidity generation, security auditing, refinement, ABI generation, MCP server creation, and quality evaluation), configuration files and agent prompts, compilation and validation scripts, and the complete evaluation framework with all five quality metrics. Additionally, we release an extended version of the FSM-SCG dataset augmented with our 9,000 generated contracts, their corresponding quality scores, security audit reports, and compilation results. This comprehensive release allows other researchers to replicate our experiments, validate our findings, test the pipeline on new contract specifications, and build upon our framework for future research. By providing complete transparency into our methodology and results, we enable the research community to verify the credibility of our claims and extend this work toward more robust and reliable smart contract generation systems.

\section{Empirical Evaluation}

Table~\ref{tab:overall_metrics} presents the aggregate performance statistics across all 9,000 evaluated contracts.

\begin{wraptable}{r}{0.50\textwidth}
\centering
    \vspace{-20pt}
\caption{Overall Performance Statistics (N=9,000)}
\label{tab:overall_metrics}
\begin{tabular}{@{}lr@{}}
\toprule
\textbf{Metric} & \textbf{Value} \\
\midrule
Average Composite Score & 81.54 \\
Minimum Score & 0 \\
Maximum Score & 100.0 \\
Standard Deviation & 12.87 \\
Average Processing Time (s) & 109.96 \\
Total Evaluation Time (h) & 274.9 \\
\bottomrule
\end{tabular}
\end{wraptable}

The system achieved an average composite score of 81.54 across all contracts, corresponding to a B grade on the letter scale. This performance demonstrates that the pipeline produces contracts of generally high quality, with the vast majority falling in the 70-90 range. Average processing time per contract was 109.96 seconds, translating to approximately 1.83 minutes per contract. This includes the complete pipeline: requirement parsing, initial Solidity generation, security auditing, potential refinement iterations, quality evaluation, and ground-truth comparison.

\begin{wrapfigure}[19]{r}{0.30\textwidth}
    \centering
    \vspace{-20pt}
    \includegraphics[width=\linewidth]{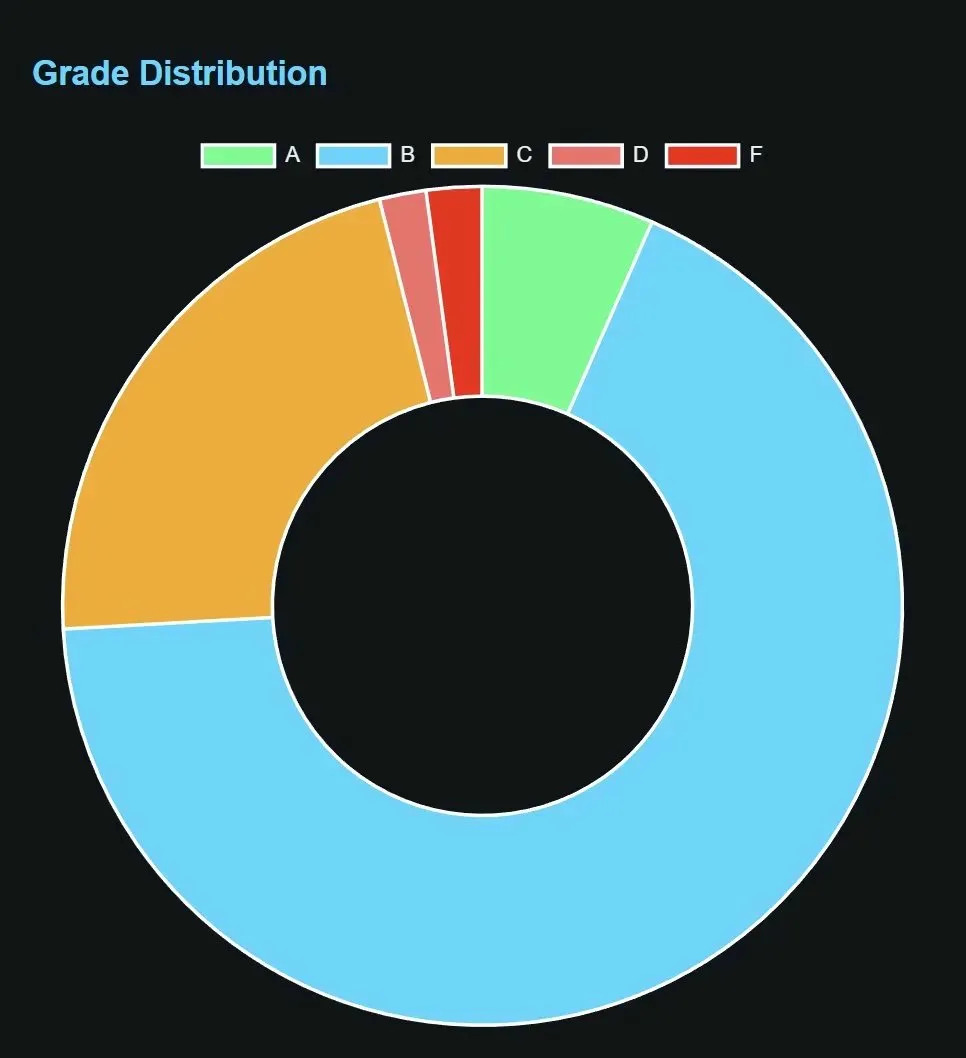}
    \vspace{-15pt}
    \caption{Grade Distribution for Generated Smart Contracts (N=9,000). The pipeline produced predominantly B-grade contracts (66.4\%), with 7.3\% achieving A-grade excellence and only 2.2\% failing completely.}
    \label{fig:grade_distribution}
\end{wrapfigure}

\subsection{Grade Distribution Analysis}

Figure~\ref{fig:grade_distribution} illustrates the distribution of letter grades across all 9,000 evaluated contracts, providing a qualitative view of system performance.

The grade distribution reveals a strong central tendency around B-grade performance (80-89 range), which accounted for 66.4\% of all generated contracts. The relatively small A-grade cohort (7.3\%) represents contracts where all five metrics achieved near-perfect scores—typically simple token transfers, basic access control contracts, or well-specified state machines with limited transitions. The low D-grade percentage (1.5\%) indicates that contracts either succeed at B-level or higher, or fail more significantly into C/F ranges, suggesting the pipeline's quality is largely determined during initial generation.

\subsection{Individual Metric Performance}

Table~\ref{tab:metric_averages} breaks down performance across the five evaluation dimensions, revealing the strengths and weaknesses of the generation pipeline.

\begin{table}[h]
\caption{Five-Dimensional Metric Averages}
\label{tab:metric_averages}
\centering
\begin{tabular}{lrrr}
\toprule
\textbf{Metric} & \textbf{Weight} & \textbf{Avg Score} & \textbf{Contribution} \\
\midrule
Functional Completeness & 25\% & 84.45 & 21.11 \\
Variable Fidelity & 15\% & 84.62 & 12.69 \\
State Machine Correctness & 15\% & 83.12 & 12.47 \\
Business Logic Fidelity & 35\% & 76.73 & 26.86 \\
Code Quality & 10\% & 83.85 & 8.39 \\
\midrule
\textbf{Composite Score} & 100\% & --- & \textbf{81.52} \\
\bottomrule
\end{tabular}
\end{table}

Variable Fidelity achieved the highest average score (84.62), reflecting strong preservation of specification terminology and naming consistency. Functional Completeness (84.45) and Code Quality (83.85) also performed well, indicating reliable implementations. Business Logic Fidelity scored lower (76.73) despite its highest weight (35\%), as it captures the most challenging semantic aspects, such as economic logic, obligations, timing constraints, and conditional flows; its weighted contribution (26.86 points) underscores its dominant impact on overall quality.

\subsection{Compilation Validation Results}

Table~\ref{tab:compilation} summarizes the compilation success rates across all batches, representing a critical quality gate for deployability.

\begin{wraptable}{r}{0.45\textwidth}
\centering
\vspace{-20pt}
\caption{Compilation Statistics Across All Batches}
\label{tab:compilation}
\begin{tabular}{@{}lr@{}}
\toprule
\textbf{Statistic} & \textbf{Value} \\
\midrule
Total Contracts Checked & 8,824 \\
Successful Compilations & 7,637 \\
Failed Compilations & 1,187 \\
Not Checked & 176 \\
\textbf{Success Rate} & \textbf{86.54\%} \\
\bottomrule
\end{tabular}
\end{wraptable}

The pipeline achieved an 86.54\% compilation success rate, meaning that approximately 6 out of every 7 generated contracts produced valid, deployable Solidity bytecode. This high success rate demonstrates that the Generator Agent reliably produces syntactically correct code adhering to Solidity 0.8.x language specifications. The compilation success rate remained remarkably consistent across all six batches (86.48\%--87.18\% range), confirming the pipeline's reproducibility.

\subsection{AI Generated vs. Expert Implementations}

The most critical evaluation component is the paired comparison between generated contracts and ground-truth expert implementations. Table~\ref{tab:ground_truth_agg} presents the aggregate statistics.

\begin{wraptable}{r}{0.50\textwidth}
\centering
\vspace{-20pt}
\caption{Aggregate Comparison}
\label{tab:ground_truth_agg}
\begin{tabular}{@{}lrr@{}}
\toprule
\textbf{Metric} & \textbf{Generated} & \textbf{Ground Truth} \\
\midrule
Avg Composite Score & 81.54 & 73.25 (+8.29) \\
Std Deviation & 12.87 & 14.52 \\
Min Score & 0 & 0 \\
Max Score & 100.0 & 100.0 \\
\bottomrule
\end{tabular}
\end{wraptable}

Strikingly, generated contracts outperformed ground-truth implementations by an average of +8.29 points. This counterintuitive result reflects the pipeline's optimization for specification-to-code fidelity rather than absolute code quality. The LLM-based Generator Agent follows prompts with extreme literalism, extracting exact function names, variable identifiers, and state labels from specifications and reproducing them verbatim. In contrast, human developers often deviate for architectural reasons, gas optimization, or clarity.

Table~\ref{tab:metric_deltas} quantifies the per-metric deltas. The largest delta occurred in State Machine Correctness (+12.84 points, +18.3\%), where generated contracts explicitly define enum-based state variables and enforce state guards in every relevant function, as mandated by the generation prompts. Ground-truth implementations often use implicit state tracking (e.g., Boolean flags, timestamp comparisons), reducing State Machine Correctness scores despite functional equivalence.

\begin{table}[h]
\caption{Per-Metric Comparison}
\label{tab:metric_deltas}
\centering
\begin{tabular}{lrrrr}
\toprule
\textbf{Metric} & \textbf{Gen.} & \textbf{GT} & \textbf{$\Delta$} & \textbf{$\Delta$\%} \\
\midrule
Functional Completeness & 84.45 & 77.82 & +6.63 & +8.5\% \\
Variable Fidelity & 84.62 & 79.13 & +5.49 & +6.9\% \\
State Machine Correctness & 83.12 & 70.28 & +12.84 & +18.3\% \\
Business Logic Fidelity & 76.73 & 66.41 & +10.32 & +15.5\% \\
Code Quality & 83.85 & 75.19 & +8.66 & +11.5\% \\
\bottomrule
\end{tabular}
\end{table}

\begin{figure}[H]
\centering
\fbox{\includegraphics[width=1\textwidth]{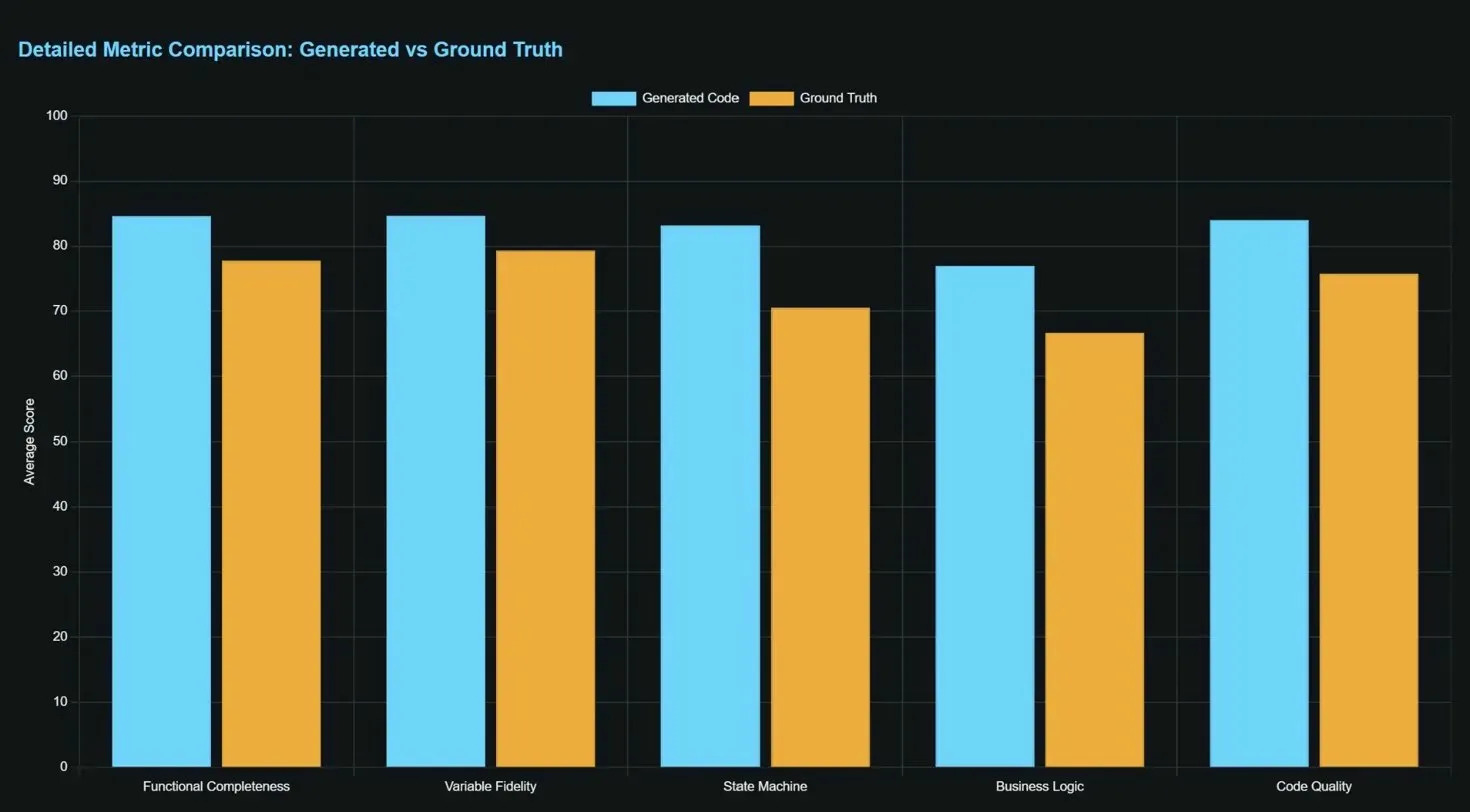}}
\caption{Detailed Metric Comparison: Generated vs. Ground Truth. Generated contracts (blue) consistently outperform ground truth (orange) across all dimensions.}
\label{fig:metric_comparison}
\end{figure}

Figure~\ref{fig:metric_comparison} visualizes the per-metric performance comparison between generated contracts and ground truth.

The distribution of score deltas (Generated minus Ground Truth) across all 9,000 contracts exhibits a pronounced positive skew, with most contracts in the +5 to +15 range. This systematic positive bias reflects the LLM's literal adherence to specification terminology—extracting exact function names, variable identifiers, and state labels verbatim—while human developers often deviate for architectural optimization or clarity. The negative tail (22.6\% of contracts) represents cases where human expertise in complex business logic, multi-party coordination, or domain-specific patterns significantly exceeded LLM capabilities. This bimodal pattern underscores the fundamental trade-off between specification fidelity and architectural quality in LLM-based contract generation.

\subsection{Error Modes and Complexity Analysis}

\begin{wraptable}{r}{0.50\textwidth}
\centering
\vspace{-20pt}
\caption{Systematic Error Modes in Lower-Performing Contracts}
\label{tab:error_modes}
\begin{tabular}{@{}lrr@{}}
\toprule
\textbf{Error Mode} & \textbf{Count} & \textbf{\%} \\
\midrule
Logic Omissions & 847 & 35.3\% \\
State Transition Errors & 562 & 23.4\% \\
Compilation Failures & 423 & 17.6\% \\
Incomplete Financial Logic & 318 & 13.3\% \\
Access Control Gaps & 248 & 10.3\% \\
\bottomrule
\end{tabular}
\end{wraptable}

Table~\ref{tab:error_modes} categorizes the primary failure types observed across 2,398 contracts in the C/D/F grade range.

Logic omissions represent the most common failure mode (35.3\%), involving missing implementation of specified obligations, conditions, or side effects. State transition errors (23.4\%) occur in contracts with complex state machines, where the system omits edge-case transitions or enforces guards inconsistently.

Table~\ref{tab:complexity_performance} shows performance stratified by specification complexity.

\begin{table}[h]
\caption{Performance by Specification Complexity}
\label{tab:complexity_performance}
\centering
\begin{tabular}{lrrr}
\toprule
\textbf{Complexity} & \textbf{N} & \textbf{Avg Score} & \textbf{Comp. Rate} \\
\midrule
Low (1-3 funcs, 1-2 states) & 3,245 & 87.2 & 94.1\% \\
Medium (4-7 funcs, 3-4 states) & 4,517 & 81.4 & 86.7\% \\
High (8+ funcs, 5+ states) & 1,229 & 71.8 & 73.2\% \\
\bottomrule
\end{tabular}
\end{table}

Performance correlates inversely with specification complexity: low-complexity contracts achieved 87.2 average score with 94.1\% compilation success, while high-complexity contracts scored 71.8 with 73.2\% compilation success. This suggests the current architecture may be insufficient for highly complex contracts.

\subsection{Security Refinement Loop Effectiveness}

\begin{wraptable}{r}{0.50\textwidth}
\centering
\vspace{-20pt}
\caption{Security Refinement Loop Effectiveness}
\label{tab:refinement_impact}
\begin{tabular}{@{}lrr@{}}
\toprule
\textbf{Metric} & \textbf{Before} & \textbf{After} \\
\midrule
Contracts w/ Med+ Severity & 4,127 & 1,203 \\
Avg Security Issues/Contract & 2.8 & 0.7 \\
Critical Vulnerabilities & 287 & 34 \\
Compilation Success Rate & 81.2\% & 86.5\% \\
\bottomrule
\end{tabular}
\end{wraptable}

Table~\ref{tab:refinement_impact} summarizes the security improvement from the audit-refine reinforcement loop.

The refinement loop reduced contracts with medium+ severity issues from 45.9\% to 13.4\%—a 70.9\% reduction. Critical vulnerabilities plummeted from 287 to 34 (88.2\% reduction), demonstrating effective iterative security improvement.

\subsection{Summary}

These findings establish that LLM-based agentic pipelines can reliably generate deployable smart contracts from natural language specifications, achieving solid B-grade performance with high compilation success rates. The combination of batch processing for large-scale evaluation and interactive single-contract mode for detailed analysis provides a comprehensive framework for both empirical research and practical application in smart contract synthesis.

\section{Conclusion}

We presented an end-to-end framework that unifies smart contract generation, verification, and comparative evaluation into a single pipeline, providing a systematic basis for empirical research on smart contract synthesis quality. The multi-stage agentic architecture, orchestrated by CrewAI-style agent teams with an iterative refinement loop, demonstrates the feasibility of automated smart contract synthesis with comprehensive quality assessment.

Key contributions include: (1) the multi-stage pipeline with compilation and security gate checks, (2) the five-dimensional quality evaluation rubric grounded in functional and semantic fidelity, (3) paired evaluation against ground-truth implementations producing per-metric deltas for targeted error analysis, and (4) reproducible benchmarking infrastructure with structured artifacts and provenance metadata.

By standardizing evaluation with transparent metrics, grounded comparisons to expert implementations, and verifiable compilation checks, this framework reduces the risk of silently deploying defective contracts and provides a practical bridge between research and deployment. It makes quality measurable, diagnosable, and auditable---thereby accelerating trustworthy adoption of generative methods in safety-critical blockchain systems.

Our evaluation of 9,000 contracts demonstrates practical viability with 81.54 average score and 86.54\% compilation success. The counterintuitive finding that generated contracts outperform ground-truth implementations by 8.29 points reveals that automated approaches excel in literal adherence to requirements while human developers optimize for architectural considerations. The security refinement loop's 70.9\% reduction in medium-severity vulnerabilities demonstrates that iterative agent-based remediation effectively bridges the gap between initial generation and production-ready security~\cite{slither2019, mythril2018}, establishing multi-agent systems~\cite{yao2023react, wei2022chain} as a promising path toward automated smart contract development.

\subsection{Limitations}

The current framework exhibits several limitations that constrain its applicability. First, dependence on LLM capabilities introduces inherent uncertainty through potential hallucinations~\cite{brown2020gpt3, openai2024gpt4}, particularly for ambiguous specifications. This manifests most severely in complex mathematical operations and cryptographic primitives, where symbolic correctness does not guarantee semantic accuracy, requiring manual verification by domain experts for production deployment~\cite{atzei2017survey, chen2020survey}.

Second, while the five-dimensional rubric provides comprehensive coverage, it cannot capture all aspects of contract correctness. Notably absent are metrics for gas optimization~\cite{wood2014ethereum}, which directly impacts deployment costs and transaction fees. A contract may achieve perfect scores while consuming unnecessarily high gas, rendering it economically impractical.

Third, performance analysis reveals strong inverse correlation between specification complexity and generation quality. High-complexity contracts (8+ functions, 5+ states) achieved only 71.8 average score with 73.2\% compilation success, compared to 87.2 and 94.1\% for low-complexity contracts. This suggests fundamental architectural limitations in single-pass generation. The 35.3\% prevalence of logic omissions in lower-performing contracts indicates current LLMs struggle to maintain semantic coherence across lengthy specifications with multiple interacting obligations.

Fourth, the evaluation framework itself carries methodological limitations. Ground-truth comparisons rely on human-authored implementations that may contain bugs or suboptimal designs~\cite{luu2016making}. Our finding that generated contracts outperform ground truth by 8.29 points reflects adherence to literal specifications rather than absolute quality, highlighting potential evaluation bias. The FSM-SCG dataset~\cite{luo2025fsmscg} may not fully represent real-world complexity involving cross-contract interactions or oracle dependencies.

\subsection{Future Work}

The framework's modular architecture enables several high-priority extensions. Property-based testing integration~\cite{fsm2020smart} represents the most immediate opportunity by automatically generating test suites that verify contract invariants across diverse input spaces. Tools like Echidna and Foundry's fuzzing capabilities could integrate into the refinement loop, with detected violations feeding back to the Refiner Agent for targeted remediation, transforming evaluation into continuous verification.

Domain-specific compliance checking represents a critical gap between technical correctness and legal validity~\cite{capponi2023defi}. Different regulatory frameworks impose distinct requirements: securities tokens must enforce transfer restrictions, DeFi protocols must implement sanctions screening, and real-estate tokenization requires jurisdiction-specific rules. A compliance extension would maintain regulatory patterns as code templates, with a Compliance Agent checking generated contracts against applicable profiles.

Gas optimization remains absent from current evaluation despite its importance to deployment economics~\cite{wood2014ethereum}. A dedicated Gas Optimizer Agent could analyze contracts for common inefficiencies—storage layout optimization, function visibility refinement, loop unrolling, and redundant SLOAD elimination—producing gas-optimized variants with functional equivalence.

Finally, fine-tuning specialized models on smart contract corpora offers a path toward improving domain-specific generation quality~\cite{roziere2024code, chen2021evaluating}. Models trained on verified contracts, audit reports, and vulnerability databases should exhibit superior understanding of blockchain-specific patterns. The FSM-SCG dataset~\cite{luo2025fsmscg} combined with real-world contracts from Etherscan and security advisories would create a comprehensive training corpus, potentially achieving higher baseline quality while reducing refinement iterations required for production readiness.

\section*{Acknowledgments}

We thank the IBM Agentics team for framework support and the open-source community for the datasets enabling this research.

\bibliographystyle{unsrtnat}
\bibliography{references}

\appendix

\section{Phase 1: Requirement Specification Agent}

The schema captures:

\begin{itemize}
    \item \textbf{Parties}: Names, roles, and addresses of contract participants.
    \item \textbf{Financial Terms}: Amounts, currencies, purposes, frequencies, and due dates.
    \item \textbf{Dates}: Contract dates, deadlines, and time-based conditions.
    \item \textbf{Assets}: Types, descriptions, locations, and values of involved assets.
    \item \textbf{Obligations}: Party responsibilities with deadlines and breach penalties.
    \item \textbf{Conditions}: Function names, variable names, state names, transitions, events, and logic conditions.
    \item \textbf{Termination Conditions}: Criteria for contract termination.
\end{itemize}

\section{Phase 2: Solidity Generation Agent}

\subsection{Critical Generation Rules}

The generation agent operates under twelve critical rules that ensure specifications translate into enforceable on-chain logic. Every guarantee in the specification must be translated into verifiable code execution, and system-wide invariants must be maintained across all functions. Domain-specific terminology carries semantic weight; for instance, the term ``Token'' implies full ERC20 standard compliance rather than a simple variable name. The agent prioritizes semantic fidelity over mere syntactic name matching, ensuring that generated code captures the intended business logic rather than surface-level terminology.

State machines must be explicitly enforced through enum-based state variables and transition guards, while access control mechanisms require explicit justification and enforcement in the implementation. The agent prohibits silent failures and symbolic logic that appears functional but lacks actual enforcement. Economic logic must be complete and conservative, protecting against value loss or exploitation. Time-based conditions that affect contract behavior must be properly integrated, and events must semantically match completed actions rather than serving as decorative logging. All generated code must serve a concrete purpose with no unused or decorative elements, and the implementation must maintain internal coherence while being robust against adversarial interactions.

\subsection{Forbidden Patterns}

The agent explicitly avoids empty or stub function bodies, unused state variables, silent failures (\texttt{if (condition) return;}), decorative events without state changes, and states named after operations rather than phases.

\section{Security Auditing Agent}

The Auditor Agent performs systematic security analysis to identify vulnerabilities before deployment. The audit examines eight critical categories that represent the most common and dangerous smart contract vulnerabilities. Reentrancy attacks are identified by examining external calls followed by state changes and violations of the Checks-Effects-Interactions pattern. Access control flaws are detected by checking for missing modifiers on critical functions and proper constructor initialization. Arithmetic safety issues include unchecked operations, proper SafeMath usage, and division by zero vulnerabilities.

The agent also evaluates ether handling security, including payable function access control, withdrawal validation, and prevention of locked ether. Denial-of-service vulnerabilities are identified through unbounded loops and external calls within loops. Input validation checks ensure proper require statements and zero address checks are present. Timestamp dependence issues are flagged when critical logic relies on block timestamps, which can be manipulated by miners. Finally, external call safety is verified by checking return value handling and error handling mechanisms. Each identified vulnerability is assigned a severity rating (none, low, medium, high, or critical) based on exploitability and potential impact, with specific line references and concrete remediation steps provided rather than generic advice.

\section{Quality Evaluator Agent}

The Quality Evaluator Agent assesses generated contracts using a five-dimensional framework with weighted scoring. Each dimension captures a distinct aspect of contract quality, as shown in Table~\ref{tab:evaluation_metrics}.

\begin{table}[h]
\caption{Smart Contract Evaluation Metrics}
\label{tab:evaluation_metrics}
\centering
\begin{tabular}{llp{8cm}}
\toprule
\textbf{Metric} & \textbf{Wt.} & \textbf{Description} \\
\midrule
\textbf{M1: Functional} & 25\% & Function name matching (exact and semantic); implementation\\
\textbf{Completeness} & & quality (logic, access control, events) \\
\midrule
\textbf{M2: Variable/} & 15\% & State variable completeness and \\
\textbf{Parameter Fidelity} & & types; variables actively used; \\
& & parameter accuracy \\
\midrule
\textbf{M3: State Machine} & 15\% & All states defined; valid \\
\textbf{Correctness} & & transitions; state guards enforced \\
\midrule
\textbf{M4: Business} & 35\% & \textit{Most heavily weighted:} \\
\textbf{Logic Fidelity} & & obligations, financial logic, \\
& & temporal logic, conditionals \\
\midrule
\textbf{M5: Code Quality} & 10\% & No placeholders/TODOs; error \\
& & messages; events; structure \\
\midrule
\textbf{Total} & 100\% & \\
\bottomrule
\end{tabular}
\end{table}

\section{Finite State Machines in Smart Contracts}

Finite State Machines (FSMs) provide a formal foundation for modeling smart contract behavior. An FSM representation specifies \textit{states} as discrete phases of contract lifecycle (e.g., Active, Pending, Completed), \textit{transitions} as conditions and actions triggering state changes, \textit{guards} as preconditions that must hold for transitions to occur, and \textit{actions} as operations executed during transitions.

\section{System Components}

The implementation consists of several core modules:

\subsection{Core Package Structure}

\begin{itemize}
    \item \texttt{agents.py}: Agent creation and configuration for CrewAI
    \item \texttt{translator.py}: Main orchestrator class coordinating all phases
    \item \texttt{programs.py}: Legacy IBM Agentics Program classes for backward compatibility
    \item \texttt{schemas.py}: Pydantic data models for contract representation
    \item \texttt{task\_builders.py}: Comprehensive prompt templates for each agent
    \item \texttt{solidity\_compiler.py}: Compilation checking using solcx
\end{itemize}

\subsection{Agent Definitions}

Each agent is configured with specific role, goal, and backstory:

\begin{lstlisting}[language=Python, caption=Agent Configuration Example]
parser_agent = Agent(
    role="Contract Analysis Expert",
    goal="Extract precise information from 
          legal contracts",
    backstory="You are an expert contract 
               analyst specializing in 
               extracting exact terminology...",
    llm=crew_llm,
    verbose=False,
    allow_delegation=False
)
\end{lstlisting}

\section{Demo Application}

The \texttt{launch\_demo.py} script provides an interactive demonstration environment for exploring the contract generation pipeline. The script orchestrates multiple services that enable both dataset exploration and real-time contract translation testing. An HTTP server is started on port 8000 to serve the web interface components, while a Flask API runs on port 5000 to handle translation requests from the frontend. The system automatically opens two browser interfaces: \texttt{sampler.html} for browsing the dataset of contract specifications, and \texttt{demo.html} for interactive testing of the translation pipeline.

Users can browse through the dataset to explore different contract types and complexity levels, select specific contracts for analysis, and observe the complete translation pipeline in action with real-time quality evaluation feedback. This interactive mode complements the batch processing capabilities by providing detailed visibility into each pipeline stage, making it particularly valuable for understanding system behavior and debugging generation issues.

\section{Configuration Options}

The translator supports configuration for \texttt{model} (LLM selection, default: gpt-4o-mini), \texttt{enable\_reinforcement} (toggle security refinement loop), \texttt{enable\_deployment} (deploy to testnet), and \texttt{max\_refinement\_iterations} (maximum audit-refine cycles).

\section{Agent Prompts}

This appendix documents the complete prompts used by each agent in the pipeline.

\subsection{Parser Agent System Prompt}

\begin{lstlisting}[breaklines=true, basicstyle=\scriptsize\ttfamily]
You are an expert contract analyst who extracts EXACT, SPECIFIC information from contracts.
CRITICAL INSTRUCTIONS:
1. Extract the EXACT function names mentioned in the contract (e.g., "initializeLease", "payRent", "confirmDelivery")
2. Extract the EXACT variable names mentioned (e.g., "monthlyRent", "securityDeposit", "deliveryDate")
3. Extract the EXACT state names mentioned (e.g., "Pending", "Active", "Completed", "Terminated")
4. DO NOT use generic placeholders - use the specific terminology from the contract
5. Capture ALL conditions, transitions, and logic flows mentioned
Your goal: Create a structured representation that preserves ALL specific details from the contract text.
\end{lstlisting}

\subsection{Generator Agent System Prompt}

\begin{lstlisting}[breaklines=true, basicstyle=\scriptsize\ttfamily]
You are a Solidity expert who generates COMPLETE, FUNCTIONAL smart contracts.

Follow ALL instructions in the prompt carefully, including:
- Complete Phase 1 semantic analysis before writing code
- Follow all 12 critical generation rules
- Avoid all forbidden patterns
- Follow correct implementation patterns
- Complete all checklist items before finalizing

Return ONLY complete, production-ready Solidity code with NO placeholders.
\end{lstlisting}

\subsection{Auditor Agent System Prompt}

\begin{lstlisting}[breaklines=true, basicstyle=\scriptsize\ttfamily]
You are a blockchain security expert who audits smart contracts for vulnerabilities. You check for reentrancy, access control issues, integer overflow, and other common exploits. You provide severity ratings (none/low/medium/high/critical) based on exploitability and impact. You give specific line references and concrete remediation steps, not generic advice.
\end{lstlisting}

\subsection{Refiner Agent System Prompt}

\begin{lstlisting}[breaklines=true, basicstyle=\scriptsize\ttfamily]
You are a Solidity security specialist who fixes smart contract vulnerabilities. Given a contract and a list of security issues from an audit, you rewrite the code to address every vulnerability while maintaining the original functionality. You follow the Checks-Effects-Interactions pattern, add reentrancy guards where needed, implement proper access control, validate all inputs with require(), and ensure no silent failures. You return ONLY the fixed Solidity code.
\end{lstlisting}

\subsection{Quality Evaluator Agent System Prompt}

\begin{lstlisting}[breaklines=true, basicstyle=\scriptsize\ttfamily]
You are an expert smart contract quality analyst specializing in evaluating how well generated Solidity code implements natural language contract specifications. You systematically assess functional completeness, variable fidelity, state machine correctness, business logic implementation, and code quality. You provide detailed scoring with specific evidence from the code and specification, identifying what was implemented correctly and what is missing.
\end{lstlisting}

\subsection{MCP Server Generator Agent System Prompt}

\begin{lstlisting}[breaklines=true, basicstyle=\scriptsize\ttfamily]
You are an expert Python developer specializing in Web3.py and MCP server generation. You create complete, self-contained MCP servers with proper error handling and transaction management for smart contract interaction.

The MCP Agent generates a complete Model Context Protocol server for blockchain interaction:

mcp = FastMCP("ContractName")

@mcp.tool()
def transfer(to: str, amount: int):
    txn = contract.functions.transfer(
        Web3.to_checksum_address(to), 
        amount
    ).buildTransaction({...})
    # Sign and send transaction
    return {"tx_hash": tx_hash.hex()}
\end{lstlisting}

\section{Solidity Generation Prompt Structure}

The complete Solidity generation prompt follows a comprehensive eight-phase structure designed to ensure systematic and thorough contract implementation. The process begins with Phase 1, semantic analysis, where the agent identifies the contract's core purpose, workflow structure, implied state transitions, and economic flow patterns. This phase includes contract type detection and invariant identification to establish the foundational understanding required for correct implementation.

Phase 2 introduces the twelve mandatory critical generation rules that govern correct implementation, followed by Phase 3, which explicitly defines forbidden patterns that must be avoided. Phase 4 provides specific guidance on state machine anti-patterns to ensure proper state modeling. Phase 5 presents correct implementation patterns through concrete code examples demonstrating best practices.

The latter phases focus on applying these principles to the specific contract at hand. Phase 6 involves detailed parsing of the contract schema to extract all relevant specification details. Phase 7 implements a mandatory implementation checklist that must be completed before finalizing the code, serving as a quality gate to verify all requirements have been addressed. Finally, Phase 8 provides a contract structure template that ensures consistency in code organization and adherence to established conventions across all generated contracts.

\section{Quality Evaluation Metrics Detail}

\subsection{Metric Weights Rationale}

The metric weights reflect their relative importance to contract correctness. \textbf{Business Logic (35\%)} is most critical as incorrect business logic defeats the contract's purpose. \textbf{Functional Completeness (25\%)} is weighted heavily since missing functions prevent required operations. \textbf{State Machine (15\%)} and \textbf{Variable Fidelity (15\%)} are equally weighted, as incorrect states lead to invalid contract behavior and variable preservation is essential for auditability. \textbf{Code Quality (10\%)} is important but less critical than correctness.

\subsection{Scoring Formulas}

For each metric, points are calculated based on specific criteria. Example for Functional Completeness:

\begin{equation}
    M_1 = \frac{10 \times |ExactMatch| + 7 \times |SemanticMatch|}{|Expected| \times 10} \times 50 + Q_{impl}
\end{equation}

Where $Q_{impl}$ is the implementation quality score based on logic completeness, access control, events, and validation.

\begin{table*}[h!]
\centering
\small
\begin{tabular}{|p{0.48\textwidth}|p{0.48\textwidth}|}
\hline
\multicolumn{2}{|c|}{\textbf{Example: Staking Contract Generation and Evaluation}} \\
\hline
\multicolumn{2}{|l|}{\textbf{Natural Language Specification (excerpt):}} \\
\multicolumn{2}{|p{0.96\textwidth}|}{
\textit{A staking contract where users can stake tokens during a farming period. The contract transitions through three states: Farming Not Started, Farming Ongoing, and Farming Ended. Users can stake tokens after the start time, withdraw staked tokens, and claim rewards based on their staking period. Rewards cease when the end time arrives.}
} \\
\hline
\textbf{Generated Solidity Code (simplified)} & \textbf{Evaluation Results} \\
\hline
\begin{minipage}[t]{0.46\textwidth}
\vspace{2pt}
\begin{lstlisting}[basicstyle=\ttfamily\scriptsize, frame=none]
enum State { 
  FarmingNotStarted,
  FarmingOngoing, 
  FarmingEnded 
}
State public currentState;
IERC20 public stakeToken;
mapping(address => UserInfo) 
  public userInfos;
function stake(uint256 _amount) 
  external afterStart beforeEnd {
  require(_amount > 0);
  stakeToken.transferFrom(
    msg.sender, 
    address(this), 
    _amount
  );
  userInfos[msg.sender]
    .amountStaked += _amount;
  emit Staked(msg.sender, _amount);
}
function withdraw(uint256 _amount) 
  external nonReentrant {
  require(_amount > 0);
  userInfos[msg.sender]
    .amountStaked -= _amount;
  stakeToken.transfer(
    msg.sender, 
    _amount
  );
  emit Withdrawn(msg.sender, _amount);
}
\end{lstlisting}
\vspace{2pt}
\end{minipage}
&
\begin{minipage}[t]{0.46\textwidth}
\vspace{2pt}
\textbf{Composite Score:} 87.3/100 (Grade: B)

\vspace{4pt}
\textbf{Strengths:}
\begin{itemize}[leftmargin=*, itemsep=0pt, parsep=2pt]
\item All required functions (stake, withdraw, claim, totalValue) implemented with complete logic
\item Explicit state machine with proper enum definition and transition guards
\item Strong access control with modifiers (afterStart, beforeEnd, nonReentrant)
\item Proper event emissions for all state changes
\item Comprehensive input validation
\end{itemize}

\vspace{4pt}
\textbf{Weaknesses:}
\begin{itemize}[leftmargin=*, itemsep=0pt, parsep=2pt]
\item Missing \texttt{poolInfos} variable from specification
\item No NatSpec documentation comments
\item Compilation error due to interface placement
\end{itemize}

\vspace{4pt}
\textbf{Metric Breakdown:}
\begin{itemize}[leftmargin=*, itemsep=0pt, parsep=2pt]
\item Functional Completeness: 92/100
\item Variable Fidelity: 85/100
\item State Machine: 90/100
\item Business Logic: 86/100
\item Code Quality: 80/100
\end{itemize}
\vspace{2pt}
\end{minipage} \\
\hline
\end{tabular}
\caption{Example of generated smart contract and automated evaluation. The specification describes a token staking mechanism with time-bounded farming periods. The generated code successfully implements the core state machine and business logic, achieving a B grade (87.3/100). The evaluation identifies both strengths (complete function implementation, proper state guards) and weaknesses (missing variable, lack of documentation), demonstrating the framework's ability to provide actionable feedback.}
\label{tab:example_evaluation}
\end{table*}

\end{document}